\documentclass{svproc}
\usepackage{graphicx}
\usepackage{nicefrac}
\usepackage{comment}
\usepackage{graphicx}
\usepackage{subfigure}
\usepackage{eurosym}
\usepackage{booktabs}
\usepackage{multirow} 
\usepackage{hyperref}
\usepackage{everypage}

\usepackage{url}

\usepackage{color}

\begin{document}
\mainmatter              
\title{Low Cost Machine Vision for Insect Classification}
\titlerunning{Machine Vision for Insect Classification}  

\author{Danja Brandt \inst{1} and Martin Tschaikner \inst{1}, \\ Teodor Chiaburu \inst{1},
Henning Schmidt \inst{1},  
Ilona Schrimpf \inst{2}, Alexandra Stadel \inst{2} and Ingeborg E.~Beckers \inst{1}, Frank Hau{\ss}er \inst{1}}
\authorrunning{Danja Brandt et al.} 
%
\tocauthor{Danja Brandt, Martin Tschaikner, Teodor Chiaburu, Henning Schmidt, Ilona Schrimpf, Alexandra Stadel, Ingeborg Beckers, Frank Hau{\ss}er}
\institute{Berliner Hochschule fuer Technik,\texttt{http://bht-berlin.de} \\
\email{Danja.Brandt@bht-berlin.de}
\and
UBZ Listhof e.V., Reutlingen, \texttt{http://www.listhof-reutlingen.de}.
}

\maketitle              
\begin{abstract}

Preserving the number and diversity of insects is one of our society's most important goals in the area of environmental sustainability. A prerequisite for this is a systematic and up-scaled monitoring in order to detect correlations and identify countermeasures. Therefore, automatized monitoring using live traps is important, but so far there is no system that provides image data of sufficient detailed information for entomological classification.

In this work, we present an imaging method as part of a multisensor system developed as a low-cost, scalable, open-source system that is adaptable to classical trap types. The image quality meets the requirements needed for classification in the taxonomic tree. Therefore, illumination and resolution have been optimized and motion artefacts have been suppressed. The system is evaluated exemplarily on a dataset consisting of 16 insect species of the same as well as different genus, family and order. We demonstrate that standard CNN-architectures like ResNet50 (pretrained on iNaturalist data) or MobileNet perform very well for the prediction task after re-training. Smaller custom made CNNs also lead to promising results. Classification accuracy of $>96\%$ has been achieved. Moreover, it was proved that image cropping of insects is necessary for classification of species with high inter-class similarity.

\end{abstract}
\section{Introduction}
\label{s:intro}

With over 1 million species worldwide, insects are the largest group of animals on our planet. Many ecological and economic interactions of insects with their environment are still unexplored. The importance of insects for humans is becoming increasingly clear. One example is their pollination service, which is estimated at 3.8 billion \texteuro \, for Germany alone\cite{LIPPERT2021106860}. All over the world the number of insect species and individuals is declining sharply \cite{sanchez2019worldwide}. In Germany, the Krefeld study reports a 75\% decrease in insect biomass from 1989 to 2015 \cite{Hallmann2017}. 
Both in the short term and in the long term, knowledge about the occurrence, distribution and population trends of insects in correlation with environmental influences is of great importance. 
Still most classical monitoring systems consist of dead traps. Biomass is determined and only a few species are identified.
In order to avoid blind spots in monitoring 
and to multiply expert knowledge, research in the field of automated classification systems is becoming more and more important. 
In this context, three partly overlapping use cases can be identified: 

\begin{enumerate}
\item \textbf{Mobile applications} (Picture Insect, Insect Identifier,  a.o. ) \textbf{and Open Source Citizen Science applications}, which contribute to raising public awareness through gamification approaches. Large amounts of data are generated this way. In particular, the iNaturalist platform serves as a global database and had an enormous impact in scientific research questions \cite{van2018inaturalist}. It also serves as a networking tool. 
Large data sets of images are labeled by the community in a collaborative manner (whenever more than two thirds of the annotators agree on the label, the sample is awarded a \textit{Research Grade} as a seal of quality). However, these randomly collected cell phone or camera images cannot properly contribute to scientific monitoring which rather requires standardized data 
collected systematically over a period of time at fixed locations. 
\item \textbf{Monitoring for pest control and identification of beneficial insects in agriculture} \cite{lima2020automatic}. In this use case, multilabel classification machine learning algorithms are used to identify certain insects species, which are of special interest \cite{wu2019ip102,Kirkeby2021}.

\item \textbf{Biodiversity monitoring} \cite{vanKlink2022}
Systematic monitoring of insect diversity requires suitable sensor systems and multilabel classification algorithms for fine-grained images or multisensor data. Moreover, the data sets are highly unbalanced due to rare insect species.
\end{enumerate}

Automated monitoring uses different technologies for insect species detection and differentiation \cite{vanKlink2022}. The data acquisition for monitoring is mainly based on wingbeat frequency measurements and image data. For wing beat frequencies acquisition alone a wide variety of different methods is used, including acoustic sensors \cite{Mankin2021}, radar \cite{rhodes2022recent,Wang2017}, multispectral analysis of reflected light \cite{Rigakis2019,rydhmer2022automating}, as well as capacitance change \cite{Khoo2020} and optoacoustic sensors \cite{Moraes2018,Wang2020,Balla2020,Potamitis2018,Batista2011}. 
There are also different approaches for imaging.
One requirement is that insects must be in relative rest for a camera to take sharp images. \cite{Amrani2022,wang2020pest24} use images from museums and other collections of dead insects for training of automated systems. \cite{Ramalingam2020,gerovichev2021high,Suto2021} use yellow traps on which the insects stick. For monitoring living insects, yellow fields~\cite{Sittinger2023} are used - suitable for pollinators -- or light traps \cite{YAO2020,hogeweg2019smart,Hoye2021} -- suitable for nocturnal insects. A recent survey of machine learning methods for insect classification is given in \cite{Teixeira_2023}.

We developed an automated low cost open source insect multisensor monitoring system where insects are counted and classified according to the taxonomic system of GBIF~\cite{GBIF} (order, family, genus, species). Citizen scientists are involved in developing and measuring. The sensor system includes an infrared optoacustic wingbeat frequency detector and a computer vision unit. For a robust classification, local environmental data are collected additionally i.e. time, temperature, humidity as well as spectral irradiance. 
While based on standard insect traps from classical entomology, our multisensor~\cite{KInsecta21} discards the container with the lethal liquid and leaves the insects unharmed.

Hence, our approach allows to deal with all three use cases as discussed above. It is designed for scientific biodiversity monitoring but due to its adaptability to various types of traps, it may as well be used to monitor specific insects e.g. by training the neural network in a one-versus-All classification.

In this paper we focus on the imaging system: the hardware setup, image post-processing as well as evaluation by solving a classification task with machine learning. A particular effort has been made to optimize and standardize image acquisition. 

The paper is structured as follows:
First, the requirements for the imaging system are summarized (2.1). Structures on which experts distinguish species need to be resolved on images of living insects in motion. According to these requirements the camera system is developed (chapter 2.2). The present dataset shows one of the biggest challenges, which stands as representative for insect monitoring: The unbalanced distribution of insect data (chapter 2.3). 
The overall system was evaluated for a data set of 16 species, some differing at the family level others only at the species level (chapter 2.4). Standard machine learning models like U-Net for semantic segmentation and ResNet-50~\cite{ResNet}, MobileNet~\cite{MobileNet} and a small self designed CNN demonstrate convincing prediction power for the insect species (Chapter 3).
Finally, results are presented and discussed in the context of currently used methods.

\section{Materials and Methods}
\label{s:materials}

\subsection{Imaging Requirements}
\label{ss:imaging}
\begin{figure}[h]
    \centering
    \includegraphics[width=0.70\textwidth]{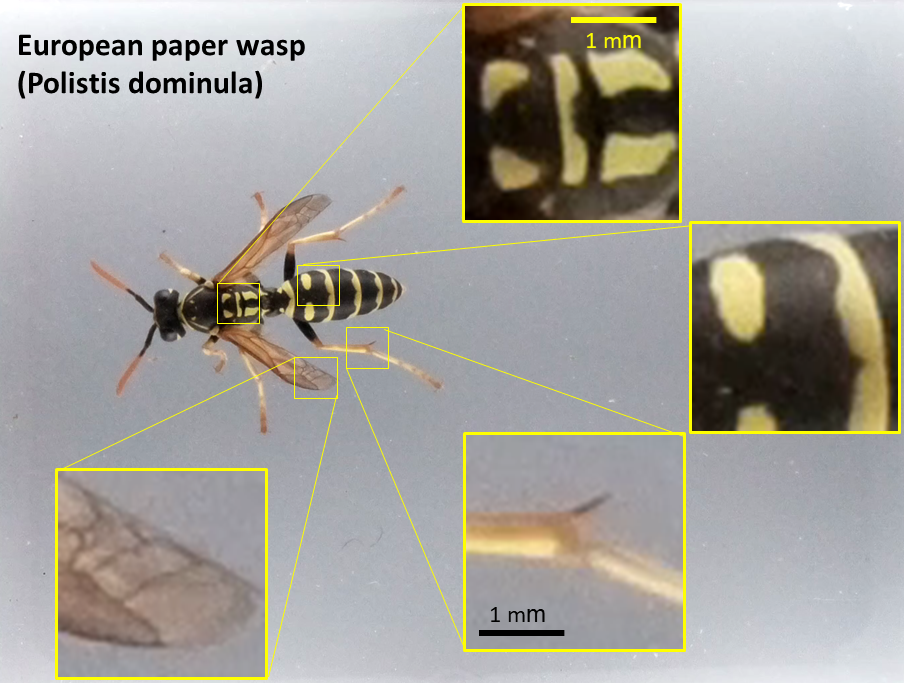}
\caption{Image of a European paper wasp taken with the Imaging Unit with zoomed-in characteristic details of body pattern, tabia-tarsus junction and wing details.}
\label{fig:waspdetails}
\end{figure}

Classifying insects from images requires taking various features into consideration such as color, contrast, patterning, size, and shape relationships.
Special features for entomological differentiation such as pubescence and wing vascular pattern may also be important for certain species, see Fig.~\ref{fig:waspdetails}. 

\begin{figure}[h]
    \subfigure{\includegraphics[width=.30\textwidth]{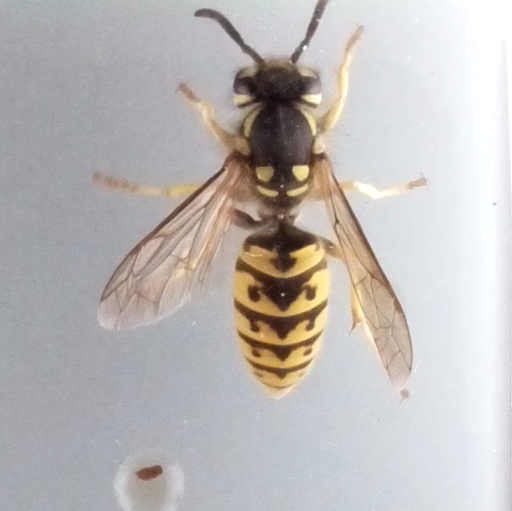}}\quad
    \subfigure{\includegraphics[width=.30\textwidth]{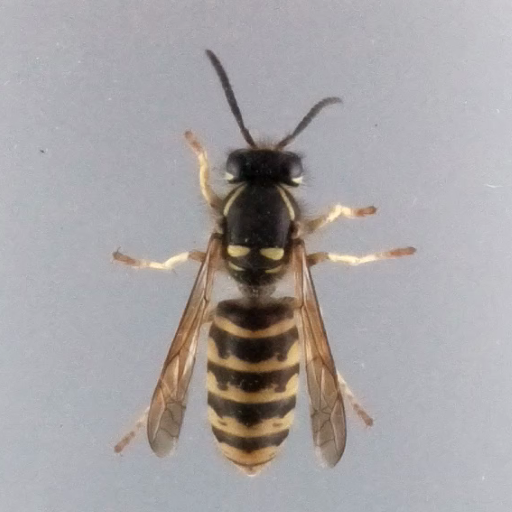}}\quad
    \subfigure{\includegraphics[width=.30\textwidth]{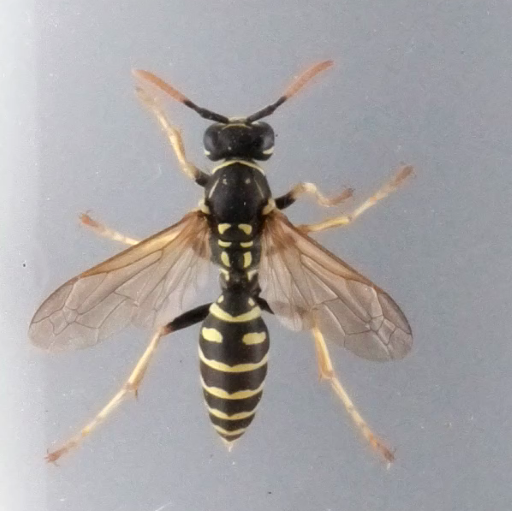}}

    \medskip
    \subfigure{\includegraphics[width=.30\textwidth]{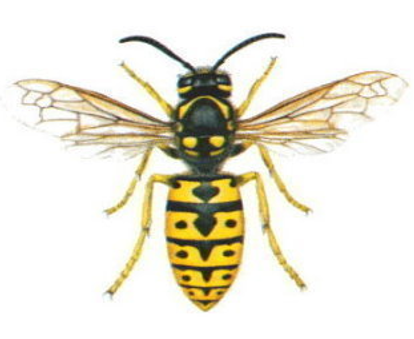}}\quad
    \subfigure{\includegraphics[width=.30\textwidth]{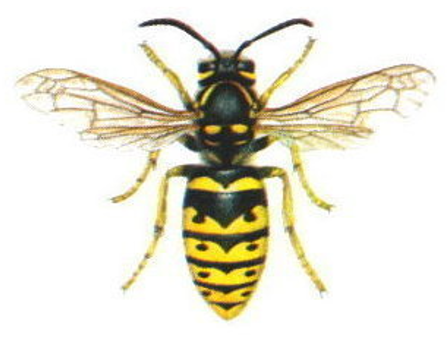}}\quad
    \subfigure{\includegraphics[width=.30\textwidth]{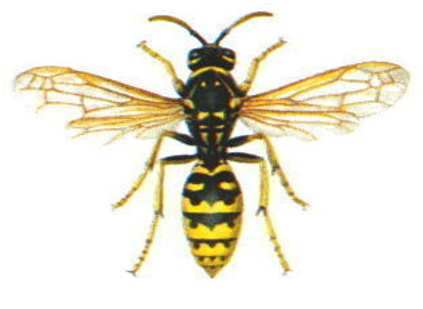}}
\caption{Comparison of 3 wasp species. (Top) Images acquired from the imaging unit. (Bottom) Schematic representation according to Ripberger \cite{koch1993ripberger}. From left to right: Vespula germanica, Vespa vulgaris, Polistes dominula}
\label{fig:images}
\end{figure}

For example, Vespula vulgaris and Vespula germanica, both of the Vespidae family, look similar but are distinguished by a particular pattern on the hind limb. On the other hand, Polistes dominula, also from the Vespidae family, has a distinctive shape relationship (Fig.~\ref{fig:images}).

The developed system is standardized according to homogeneous illumination, suppression of motion blur, resolution and depth-of-field, what is critical for an accurate classification. A homogeneous, diffuse illumination is essential as well as a resolution that matches insect structural sizes. Furthermore, shadows on the background make automatic segmentation and detection difficult. To address these requirements, the acquisition unit has diffusing panels on the side walls and the bottom. A trade-off between maximal resolving power and large depth-of-field has been figured out. Characteristic structures like the hairs or the wing veins are in the range of 20 - 100 microns. However, the distance between the veins is at least one order of magnitude greater. The choice of an effective aperture (f$/$8) leads to a resolution of 10 micrometers, which makes it possible to differentiate vein sequences on the object side. The related depth of field is 15 mm for a rear distance of 150 mm (Fig.~\ref{fig:imagingunit}). 

\subsection{Hardware Setup}

\begin{figure}[h!]
    \subfigure{\includegraphics[width=0.5\textwidth]{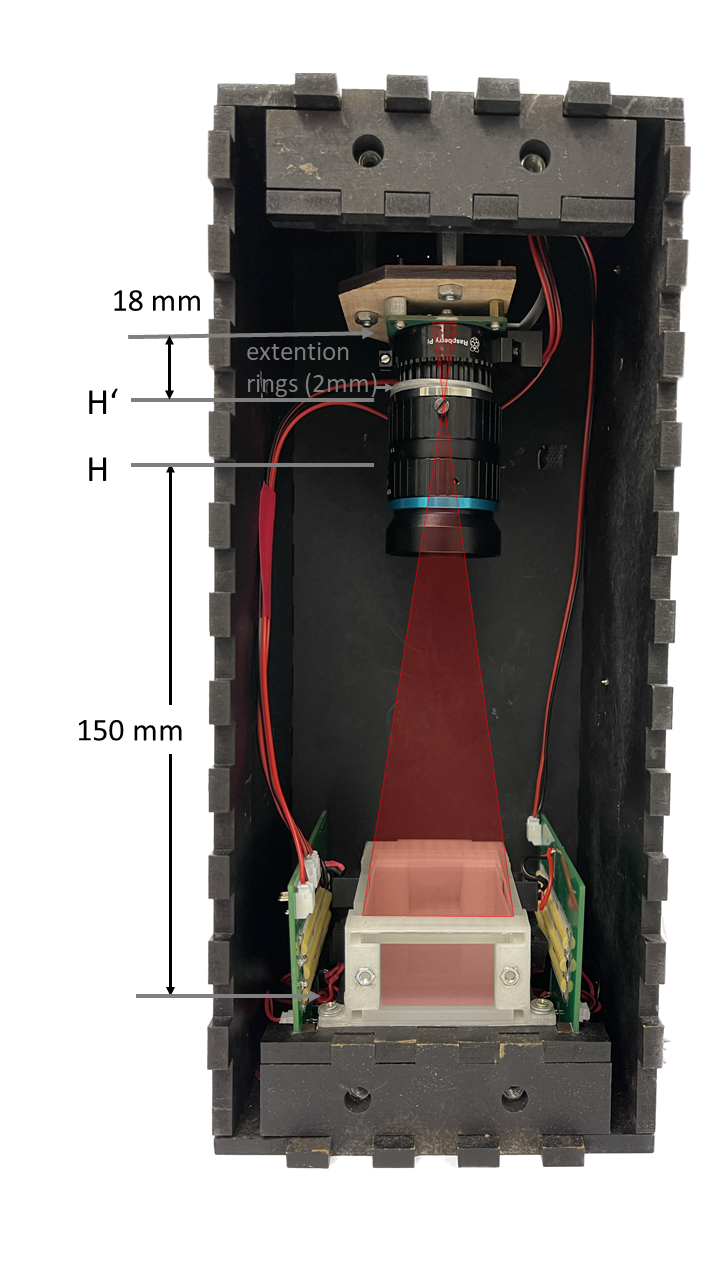}}
    \centering
    \subfigure{   
0    \includegraphics[width=0.48\textwidth]{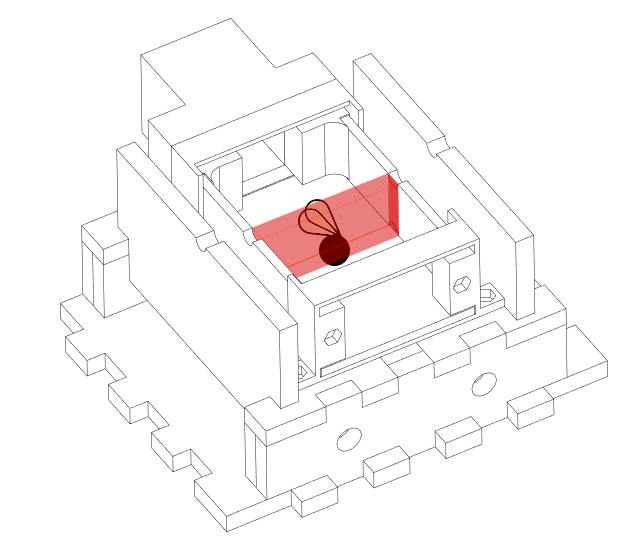}}
\caption{Image acquisition unit (left) and principle of the insect arena (right). The red region indicates the light barrier in the center of the arena that starts the image recording.}
\label{fig:imagingunit}
\end{figure}

Behind the insect trap, the multisensor system \cite{KInsecta21} 
with the imaging unit  (Fig.~\ref{fig:imagingunit} (left)) replaces the container filled with poisonous agent. This imaging unit is described below.
The area for imaging as shown in Fig.~\ref{fig:imagingunit} (right) is confined to $60 \times 45 \times 20 \,\mathrm{mm}^{3}$. The field of view (FoV) is optimized for $\mathrm{FoV} = 60  \times 45 \,\mathrm{mm}^{2}$. The adapters at the input and output are 3D printed from PLA. The arena itself is cut from Plexiglas by a laser cutter. The top side is transparent, while the side walls and bottom are made of diffusing PMMA with a light transmission of $62\,\%$. Two photoelectric sensors (IS471FE OPIC optical integrated circuit), one at the bottom and one at the top, facing each other in the center of the arena, trigger the photo detection. Insects passing by trigger the strobe flash light, consisting of three 12V LED strips with three units each (nine in total), which homogeneously illuminate the insects over the entire height. Moreover, the LED strips provide a neutral gray background by allowing light to enter the floor from the side facets. The IS471FE chip as receiver of the light barrier provides a modulated signal to an external IR LED (transmitter with $940\ \textrm{nm}$), so that interfering influences, such as ambient light, are filtered. To reduce stray light, an additional  $0.75\ \textrm{mm}$ slit aperture is applied. This increases the sensitivity of the photoelectric light barrier sensor especially for small insects such as fruit flies and hover flies. 

Custom electronic circuit boards are designed and developed to be plugged into the Raspberry Pi minicomputer. For imaging, a Raspberry Pi HQ camera (Sony CMOS sensor IMX477R, 12.3 megapixels) is used in combination with a 10MP 16mm telephoto lens with variable aperture. The entire system is located in a dark case of size $296 \times 127 \times 115\, \mathrm{mm}^3$. The length is chosen according to an optimal trade-off for resolution, field-of-view and depth of field. As a result, an additional $2\, \mathrm{mm}$ spacer ring was 3D-printed to extend the image distance, resulting in greater magnification and shorter object distance.

Flashing in a dark scene minimizes motion blur caused by CMOS rolling shutter technology. The FSTROBE pin of the camera chip sends a PWM signal synchronized with the frame rate. The combination of a short flash of 500 microseconds and an exposure time of 23.5 milliseconds, which is related to the readout time and the reset time of a single line, results in the simulation of a global shutter that minimizes distortion effects. Ants, for example, crawl very fast with a speed of $50\, \textrm{cm/s}$. During one single flash, this corresponds to a path of $0.25\, \mathrm{mm}$. At 10x reduction in magnitude, this is equivalent to 25 micrometers on the chip (13 pixels on the Sony IMX477R sensor chip). This results in motion blur 2.5 times worse than the blur due to diffraction-limited resolving power at f/8. However, most insects are slower; besides, they also slow down in the arena. It is evident in the acquired images that the wings, whose wingbeat frequency can reach values in the kHz range, as well as individual limbs, exhibit motion blur, while the trunk is detected free of motion blur, see Fig.~\ref{fig:images}. This has been demonstrated for various insects including ants. 

To calibrate the camera system a Bokeh application (Python) has been implemented, which is running on the Raspberry Pi as a local server. The automated measurement is processed by a Python script, too. It controls the individual sensors, processes the incoming data and stores the information on SD card. The camera provides an in-memory H264 video stream based on a ring buffer to limit RAM memory usage. The size of the ring buffer corresponds to an H264 video of about 1.5 seconds in length at a resolution of $1440 \times 1080$ pixels. A sudden increase in frame brightness triggers the extraction of the first image frame from the video, followed by two others, saved as png-images.

\subsection{Dataset and Preprocessing}
\begin{figure}[h!]
    \centering 
    \includegraphics[width=0.6\textwidth]{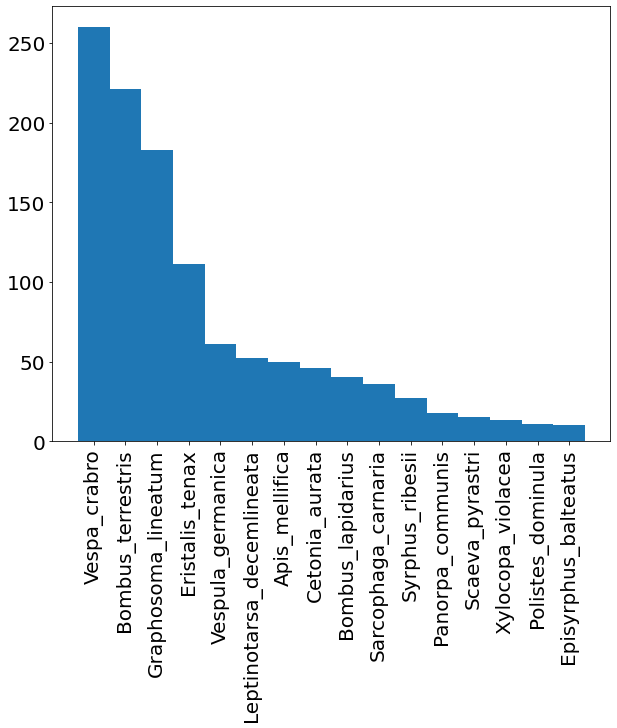}  
    \caption{Distribution of the species in the total data set used for the experiments. Notice the long tail of the histogram which is typical for insect monitoring data. 
    }
     \label{fig:LongTail}
\end{figure}

To evaluate the quality of our insect images, we will solve a  classification task using a dataset of 1154  annotated images with 16 insect species classes, which are spread over the taxonomic tree - order, family, genus, species.
The dataset is extremely unbalanced, as can be seen in Fig.~\ref{fig:LongTail}. 
This is a realistic assumption for real world data, since certain insect species will pass through the sensor system more often than others e.g. rare species. Thus, the data set represents the reality within insect monitoring very well. The detailed list of the distribution of species in the data set used in this work is given in Table~\ref{tab:species}. 

The original images are of size $1440 \times 1080$ pixels. 
We used Label Studio \cite{lstudio} to manually crop them to a square region containing the insect. 
Section~\ref{ss:semanticseg} discusses how cropping may be done automatically and very reliably using semantic segmentation.

\begin{table}[h]
    \centering
\begin{tabular}{|l|l|l|l|l|}
 \toprule
 \textbf{order} & \textbf{family} & \textbf{genus} &  \textbf{species} \\ 
\midrule
\multirow{7}{*}{Hymenoptera} & \multirow{4}{*}{Apidae} & Apis  & mellifica \\ 
 &  &  \multirow{2}{*}{Bombus} & lapidarius  \\
  &    &     & terrestris  \\
     &   & Xylocopa & violacea \\
     & \multirow{3}{*}{Vespidae} &  Vespa  & crabro \\
    &     & Vespula  &germanica  \\
   &      & Polistes & dominula  \\    
 \midrule
\multirow{2}{*}{Coleoptera} & Chrysomelidae & Leptinotarsa &  decemlineata  \\
       & Cetoniidae &  Cetonia  & aurata  \\
 \midrule 
Mecoptera & Panorpidae & Panorpa &  communis  \\
 \midrule
\multirow{5}{*}{Diptera} & Tachinidae & Sarcophaga & carnaria  \\
      & \multirow{4}{*}{Syrphidae} &  Scaeva & pyrastri  \\
     &       & Eristalis & tenax  \\
     &    &  Episyrphus  & balteatus \\
     &    & Syrphus & ribesii \\ 
 \midrule
 Heteroptera & Pentatomidae & Graphosoma  & lineatum \\
 
 \bottomrule
\end{tabular}
\vspace{3mm}

\caption{The dataset consists of 1154 images of 16 insect species from various insect orders (5), families (8) and genera (15).}
\label{tab:species}
\end{table}

\subsection{Machine Learning, Insect Classification}
This work shows that the high quality image data provided by the presented imaging system gives very good results for automatic insect classification on minicomputers like Raspberry Pis.
We decided to use two standard convolutional neural networks (CNN) of different sizes and a custom made rather small CNN. In all three cases, the input image size is $224 \times 224$.

As already stated, insect classification is a fine-grained image analysis task. Moreover, data from monitoring will always be extremly unbalanced. An additional constraint on the monitoring system is that predictions must be performed on a small devices of low power consumption, which precludes massive neural network models.

As a benchmark model, ResNet-50 \cite{ResNet} was used, which is a rather large CNN ($\approx 25$M parameters). Pretrained on the iNaturalist dataset 2021 with 2.7M images from 10k different species \cite{horn_resnet} the ResNet converged within a small number of epochs. No layers have been frozen.
Secondly, a smaller MobileNet ($\approx 3.5$M parameters, default depth parameter $\alpha = 1$) pretrained on ImageNet was retrained, also with no frozen layers. Finally, a small custom made CNN with 5 layers and $\approx 1.3M$ parameters was trained from scratch, see Fig~\ref{fig:customCNN} in the Appendix. 

The performance of the three CNNs is compared for two distinct cases: full images of the $6\, \mathrm{cm} \times 4, \mathrm{cm}$ arena, see Fig~\ref{fig:imagingunit} and cropped images of the insects. In the first case, downscaling from $1440 \times 1080$ pixels to the NN input size of $224 \times 224$ results in a serious loss of morphological details of the insects. Top-1 accuracies as well as confusion matrices are discussed in Section~\ref{ss:classification}.

The models and experiments have been implemented in Python using Keras and Tensorflow 2.8. 

\subsection{Semantic Segmentation}
\label{ss:semanticseg}
For a precise and robust object localisation, we evaluated a small sized U-Net \cite{Ronneberger2015Unet} as provided by \cite{unetMicro2021}. The images were resized from $1440\times 1080$ down to $256 \times 256$.
Label Studio \cite{lstudio} was used to segment insects on 301 images by hand. Based on this ground truth, the U-Net was trained to predict masks. Preliminary results are presented in Section~\ref{ss:segbb}. The predicted masks can be used to define a minimal bounding box containing the insect and, thus, perform an automatic cropping.

\section{Results and Discussion}

\subsection{Classification}
\label{ss:classification}
The data set was split in fixed training, validation and test sets of ratios 60/20/20. Top-1 accuracies on the test set are reported in Table~\ref{tab:accucies}  \footnote{As a side note, we did not fine-tune class weights or over-/under-sampling techniques to deal with the underrepresented classes. Nonetheless, the models performed very well on the cropped data.}.
\begin{table}[h!]
    \centering
\begin{tabular}{|c|c|c|}
 \toprule
 \textbf{Model} & \textbf{Accuracy (full)} & \textbf{Accuracy (cropped)} \\ 
\midrule
ResNet-50 & 0.9685 & 0.9604 \\
\midrule
MobileNet & 0.8776 & 0.9780 \\
\midrule
Custom CNN & 0.7058 & 0.8722\\ 
 \bottomrule
\end{tabular}
\vspace{3mm}
\caption{Comparison of insect classification performance (Top-1 accuracy) on the test set using full images and cropped images.}
\label{tab:accucies}
\end{table}

It is quite evident that all models perform very well on cropped images. Also worth emphasizing is that the MobileNet and the Custom CNN reach noticeably higher scores when trained on the cropped samples as opposed to the uncropped ones. This is, in contrast, not the case with the ResNet, which seems to perform equally well on the raw and zoomed-in images. At first sight, this appears counterintuitive, since downsampling full images to 224 $\times$ 224 pixels causes substantial information loss and results in rather coarse/blurry images. This indicates, that up to a certain classification accuracy within the taxonomic tree, the features expressed by high spatial frequencies (fine grained structures) are not relevant. Moreover, a reasonable explanation
for the still very high scores may lie in the favorable initialization with \textit{iNat21} weights. Hence, the ResNet starts the retraining already with prior knowledge about insects, which a MobileNet, for instance, did not inherit from the ImageNet weights. And indeed, the size of the ResNet (around 6 times more parameters than MobileNet and 20 times more than the custom CNN) is also not to be neglected.

\begin{figure}[h!]
    \centering 
    \includegraphics[width=0.99\textwidth]{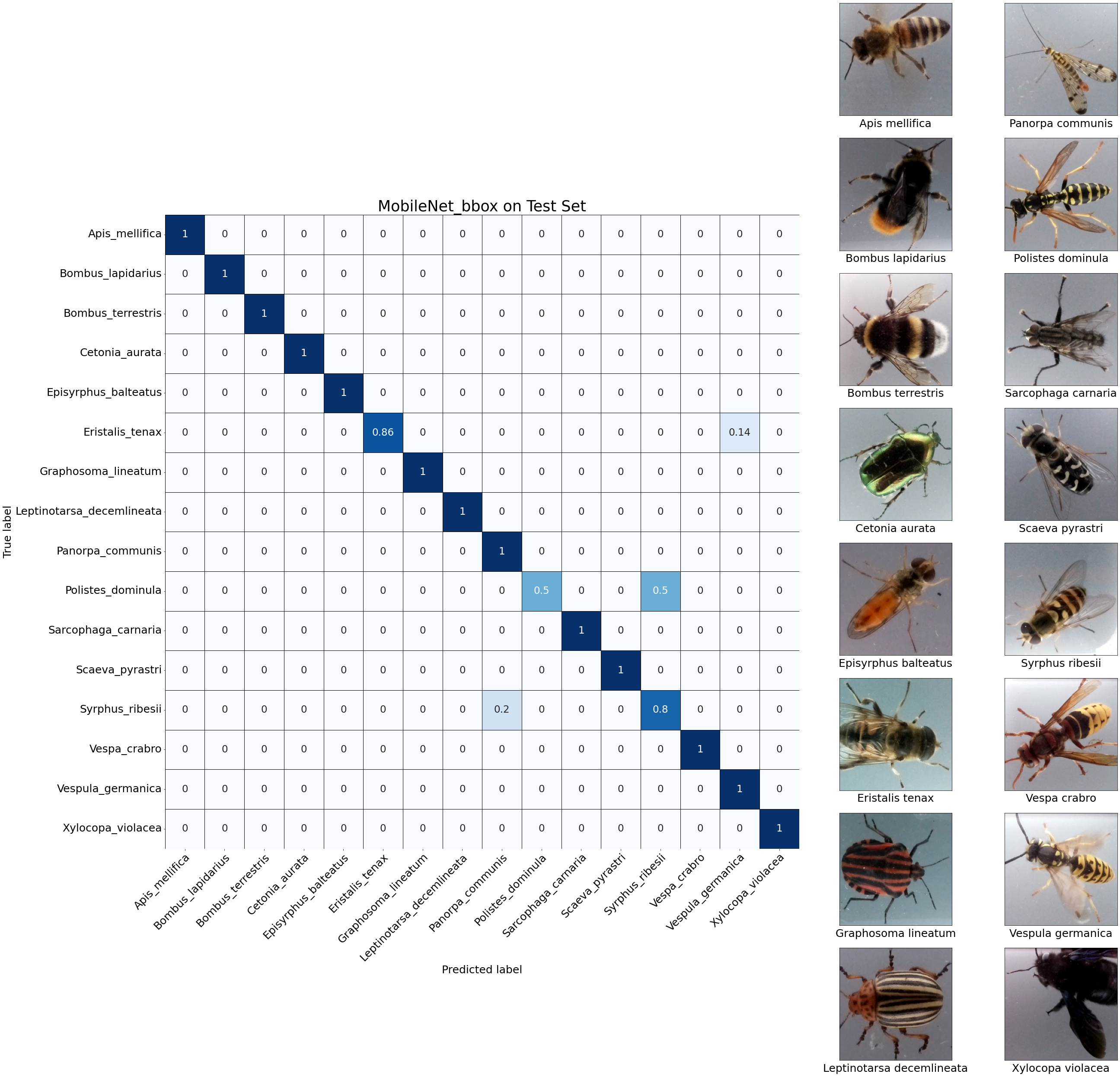}  \caption{Confusion Matrix MobileNet, test set with cropped images}
     \label{fig:confmatCropped}
\end{figure}

\begin{figure}[h!]
    \centering 
    \includegraphics[width=0.99\textwidth]{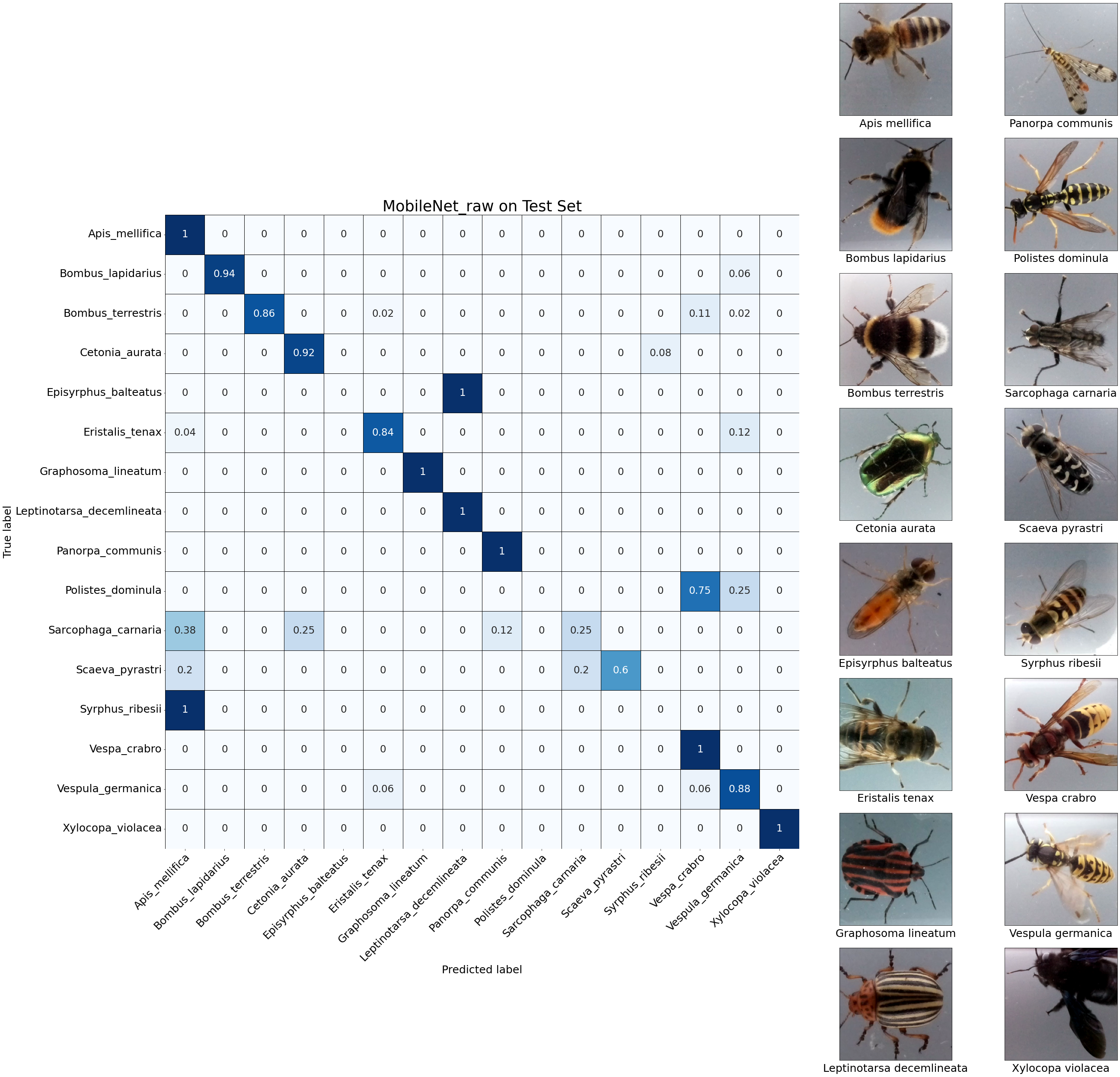}  \caption{Confusion Matrix for MobileNet, test set with full images. Although the Top-1 Accuracy (0.8776) is rather high also for full images, the classifier does not perform very well on some rare species, e.g. \textit{Polistes dominula} and \textit{Episyrphus balteatus}.}
     \label{fig:confmatFull}
\end{figure}

Since the dataset is highly unbalanced, the test accuracy alone does not provide a complete picture of the performance of the models in the individual classes. The confusion matrices in Fig.~\ref{fig:confmatFull} and Fig.~\ref{fig:confmatCropped} from the Appendix reveal that especially the rare species are considerably difficult to distinguish in full images. In particular the two rarest species \textit{Polistes dominula} and \textit{Episyrphus balteatus} (Fig.~\ref{fig:LongTail}) are never predicted correctly in the full images, see Fig.~\ref{fig:confmatFull}.

A key parameter to further improve the accuracy of the models and to handle a large number of different species is to increase the amount of data. Evidently, the more data provided from different species, the more species can be classified. For this reason, the contribution of citizen scientists in data collection is central to the project.

There are insect species for which it is very difficult or even impossible to determine the correct taxon based on multisensor data. Here, dissection or genetic tests would be necessary. Therefore, the classifier should not always predict the insect species but rather the highest possible level in the taxonomic hierarchy, e.g. species or genus, or family. To achieve this, a hierarchical classification scheme may be implemented, see Gitlab repository ~\cite{teomaster}.

\subsection{Bounding Boxes via Segmentation}
\label{ss:segbb}

\begin{figure}[h!]
    \centering
    \includegraphics[width=0.99\textwidth]{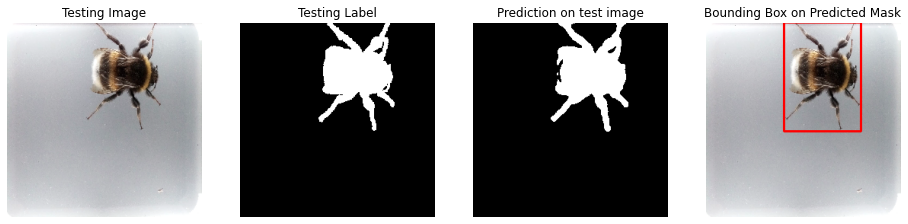}
\caption{Example of inferring a bounding box: From left to right: original image, manually annotated segmentation mask (used as label for training the U-Net in a supervised context), U-Net predicted mask, bounding box inferred from the predicted mask.} 
\label{fig:bb_semseg}
\end{figure}

The results discussed in detail in Section ~\ref{ss:classification} show that zooming in on the insect produces better results, although size information is lost when cropping. Therefore, we implement automated cropping in our classification workflow, see Fig.~\ref{fig:bb_semseg}.

The automated cropping algorithm based on segmentation and described in Section ~\ref{ss:semanticseg}, reaches an IoU of 0.7658 for the predicted masks using 301 labeled images as input. This method is more robust than commonly used thresholding techniques. Variations in brightness, white balance and reflects can occur in every image, making it impossible to set a global threshold value suitable for every image. Thresholding also leads to problems if several insects of different colours are present in one image or if certain parts of the insect have similar colours compared to the background, such as transparent wings. In addition, we observed that the arena gets dirty quickly, making it necessary to filter out non-specific objects such as dust or insect dirt from the bounding box evaluation. Prediction of insect masks overcomes these problems. 

In order to improve the preprocessing pipeline, one-stage models such as YOLO~\cite{yolo2016} or other single-shot detectors will also be tested to obtain the best bounding box evaluator suitable for our data.

\section{Conclusion and Outlook}
The goal of this project is to develop a complete and autonomous system for insect species classification for use in the field that can be replicated and operated by citizen scientists.

To achieve this, a low-cost system based on a minicomputer with various sensors and a Raspberry Pi HQ camera has been developed for standardized and AI-based automated monitoring of insects. The presented work focuses on the image acquisition unit including image processing. Both have been optimized to meet the requirements for an automated species-level classification on a minicomputer that can serve as edge device.
The image processing includes an automatic trigger to capture a video clip, an adjusted frame selection, identification of insects in each frame and cropping, and classification of the insect. 
The image processing system is evaluated based on the correlation between the preprocessing (cropping) and the classification accuracy for different CNNs (ResNet-50, MobileNet, and a Custom-defined CNN). The ResNet-50 pretrained on the iNaturalist data of insects shows the best results for full images with up to 96$\%$ accuracy, whereas the MobilNet and ResNet both achieved 96 $\%$ accuracy for cropped images. However, a custom designed network proves to be a promising approach for an energy-efficient application, in particular inference on the Raspberry Pi as edge device. In addition, we found that the accuracy of the smaller models is significantly higher trained on cropped images. 

As a next step, we will integrate multisensor data such as wing beat frequencies, environmental data and metadata such as insect size, which is known by its relationship to fixed FoV. Especially, the classification of even very small insects is sensitive to characteristic wingbeat frequencies. Prior probabilities depending on size, temperature, season and other factors, which are also used by experts, should similarly support the classification in automated monitoring. To overcome the problem of rare species, achieving high accuracy even with few input samples, class weighting and oversampling techniques will be investigated as well as methods for few shot learning such as contrastive learning. Extensive data collection with citizen scientists will be conducted during this year's flight season.

\bibliographystyle{./bibtex/spmpsci_unsrt}
\bibliography{KInsect_library}

\clearpage

\section*{Appendix}

\subsection*{A. Confusion Matrices}

$\,$

\clearpage

\subsection*{B. Custom CNN Model}

\begin{figure}[h!]
    \centering 
    \includegraphics[height=0.88\textheight]{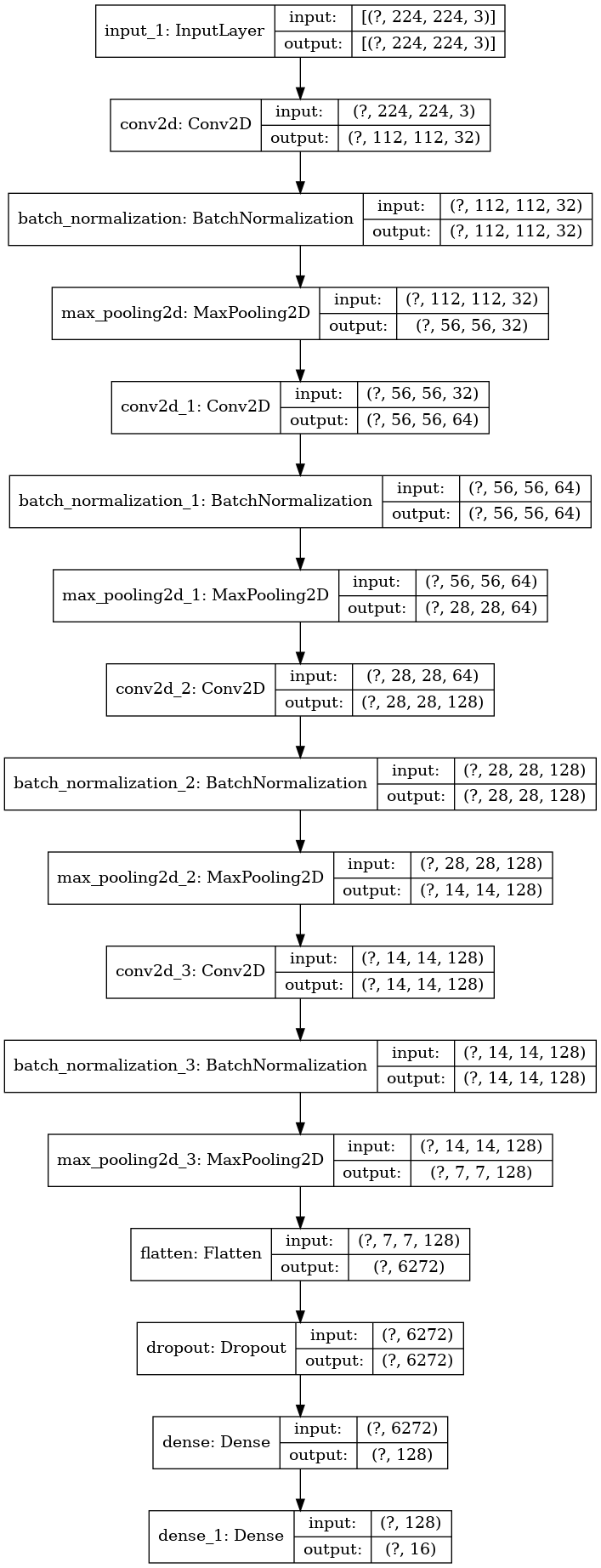}
   \caption{Custom CNN architecture - 1.270.992 parameters}
     \label{fig:customCNN}
\end{figure}

\end{document}